\documentclass[runningheads]{llncs}
\usepackage[T1]{fontenc}
\usepackage{graphicx}
\usepackage{xcolor}
\usepackage{amsmath}
\usepackage{pifont} 
\usepackage{amssymb}
\usepackage{booktabs} 
\usepackage{makecell}

\begin{document}
\title{Controllable Skin Synthesis via Lesion-Focused Vector Autoregression Model}

\author{Jiajun Sun\inst{1,2}\and
Zhen Yu\inst{3,4}\and
Siyuan Yan\inst{3,5} \and
Jason J. Ong\inst{1,2,6} \and
Zongyuan Ge\inst{3,4,}\thanks{Corresponding author.}\and
Lei Zhang \inst{1,2,7,8,*}}

\authorrunning{J. Sun et al.}
\institute{School of Translational Medicine, Faculty of Medicine, Nursing and Health Sciences, Monash University, Melbourne, Australia \and
Melbourne Sexual Health Centre, Alfred Health, Melbourne, Australia \and
AIM for Health Lab, Monash University, Melbourne, VIC, Australia \and
Faculty of IT, Monash University, Melbourne, VIC, Australia\and
Faculty of Engineering, Monash University, Melbourne, VIC, Australia\and
Faculty of Infectious and Tropical Diseases, London School of Hygiene and Tropical Medicine, London, United Kingdom \and
China-Australia Joint Research Center for Infectious Diseases, School of Public Health, Xi'an Jiaotong University Health Science Center, Xi'an, Shaanxi, China \and
Phase I clinical trial research ward, The Second Affiliated Hospital of Xi’an Jiaotong University, Xi'an, Shaanxi, China.\\
\email{\{Jiajun.Sun, Zhen.Yu1, Siyuan.Yan, Jason.Ong, Zongyuan.Ge, Lei.Zhang1\}@monash.edu}}

\maketitle  
\begin{abstract}
Skin images from real-world clinical practice are often limited, resulting in a shortage of training data for deep-learning models. While many studies have explored skin image synthesis, existing methods often generate low-quality images and lack control over the lesion's location and type. To address these limitations, we present \textbf{\textit{LF-VAR}}, a model leveraging quantified lesion measurement scores and lesion type labels to guide the clinically relevant and controllable synthesis of skin images. It enables controlled skin synthesis with specific lesion characteristics based on language prompts. We train a multiscale lesion-focused Vector Quantised Variational Auto-Encoder (VQVAE) to encode images into discrete latent representations for structured tokenization. Then, a Visual AutoRegressive (VAR) Transformer trained on tokenized representations facilitates image synthesis. Lesion measurement from the lesion region and types as conditional embeddings are integrated to enhance synthesis fidelity. Our method achieves the best overall FID score (average 0.74) among seven lesion types, improving upon the previous state-of-the-art (SOTA) by \textbf{6.3\%}. The study highlights our controllable skin synthesis model’s effectiveness in generating high-fidelity, clinically relevant synthetic skin images. Our framework code is available at https://github.com/echosun1996/LF-VAR.


\end{abstract}

\section{Introduction}

High-quality skin images are essential for training and validating medical artificial intelligence (AI) models. However, skin images from real-world clinical practice often suffer from scarcity and long-tail imbalances, leading to biased outcomes or algorithmic unfairness \cite{STAAL202152}. Furthermore, privacy concerns, legal restrictions, and financial burdens pose substantial challenges in acquiring diverse, high-quality skin images \cite{ljae040,jamadermatol.2016.6214}.

Generative models have emerged as a potential solution to address these data-related challenges. For instance, SkinDiff \cite{SkinDiff} employs a latent diffusion framework to generate labelled skin lesion images, while Derm-T2IM \cite{Derm_T2IM} leverages diffusion models with lesion-specific textual prompts to generate dermoscopic images. However, diffusion models are limited by their dependence on large-scale training data and computationally intensive synthesis processes \cite{highresolutionimagesynthesislatent,nwae348}, making them less feasible for resource-constrained clinical settings.

Recent advances in autoregressive models, particularly Vector AutoRegressive (VAR) Transformers, have demonstrated competitive performance in general image synthesis tasks while requiring fewer computational resources and less training data than diffusion models \cite{VAR}.
However, our evaluation of the VAR model in skin image synthesis revealed two critical limitations: First, existing implementations lack mechanisms to guide synthesis using diagnostically relevant features, resulting in generated images with limited pathological diversity and fidelity. Second, without explicit separation of lesions and normal skin regions, VAR-generated images exhibit unnatural transitions and background noise, reducing their realism.

To address these challenges, we propose \textit{LF-VAR}, a controllable synthesis model that leverages masks and lesion type as prompts to enable precise and flexible skin lesion synthesis. Additionally, \textit{LF-VAR} integrates quantified lesion measurements—such as shape and texture—to guide our model over clinically relevant features, ensuring high-fidelity synthetic images with well-preserved pathological characteristics. The primary contributions of our work are:

\begin{enumerate}

    \item We propose a lightweight encoder that transforms quantified lesion measurements into discriminative latent representations, which enables precise control over clinically relevant features and allows high-fidelity synthesis.
    
    \item We implement a unified model that incorporates segmentation masks and lesion type, enabling controlled synthesis of lesions across different lesion categories. This design allows for the flexible generation of diverse pathological presentations while maintaining consistency with clinical descriptors.

    \item We integrate pixel-level segmentation into the generative pipeline to focus lesion synthesis while minimizing background region artifacts. It enables high-quality image synthesis by maintaining a clear distinction between lesions and normal skin regions.

    \item We conduct extensive validation indicating robust synthesis across intra-class, inter-class, and cross-dataset scenarios. Our evaluation confirms the model’s effect in high-fidelity clinically relevant skin synthetic.

\end{enumerate}

\begin{figure}[t]
\centering
\includegraphics[width=\textwidth]{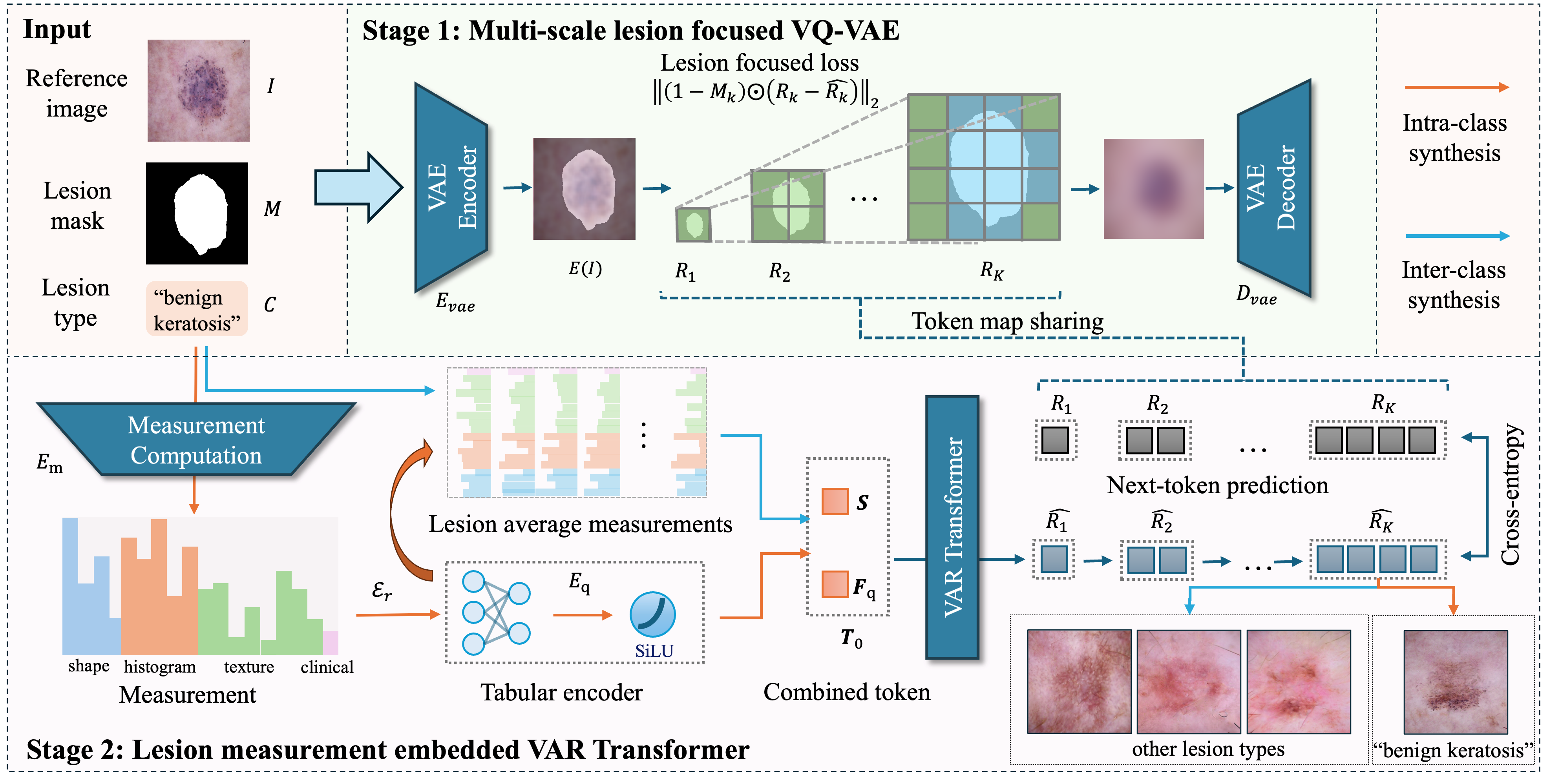}
\caption{The framework of \textit{LF-VAR}. A structured VQ-VAE encoded images into $K$ token maps, shared with the VAR, and was trained with lesion-focused loss. Lesion measurement scores were extracted from lesion areas and encoded as $\mathbf{F_q}$, combined with lesion type embedding $\mathbf{S}$, and incorporated into the VAR as a conditional token to enable controlled intra-class skin lesion synthesis. Inter-class synthesis was achieved through the VAR model by querying the lesion type average measurements codebook.} \label{figure1}
\end{figure}

\section{Method}
The proposed synthesis model, as shown in Fig.\ref{figure1} consisted of two distinct stages: (1) a multiscale lesion-focused Vector Quantised Variational Auto-Encoder (VQ-VAE) \cite{VQVAE}, which was employed to acquire compact discrete image representations, and (2) a VAR \cite{VAR} model trained on tokenised representations to facilitate image synthesis. In the second stage, measurement scores extracted from the skin lesion region were incorporated into the transformer, along with lesion types as conditional embeddings, to improve the fidelity of the generated images.

\subsection{Multi-Scale Lesion-Focused VQ-VAE}
A multi-scale lesion-focused VQ-VAE was trained to encode images into discrete latent representations, enabling structured and hierarchical tokenization. Given an input image $\mathbf{I} \in \mathbb{R}^{H \times W \times C}$, the encoder $E(\cdot)$ extracted feature maps, yielding a latent feature map $ f = E(\mathbf{I})$, where $f \in \mathbb{R}^{h \times w \times d}$. A multi-scale quantization process was subsequently applied to discretize the latent representation:
\begin{equation}
\footnotesize
    \mathbf{R} = Q(f) = \{R_k\}_{k=1}^{K}, \quad R_k \in \mathbb{Z}^{h_k \times w_k},
\end{equation}

\noindent where $Q(\cdot)$ denoted the quantization function, $K$ represented the number of scales, and each $R_k$ was a discrete token grid at scale $k$. The decoder networks $ D(\cdot) $ and $ \hat{D}(\cdot) $ were responsible for reconstructing the feature map and the final image, respectively, where the reconstructed feature map was given by $\hat{f} = D(\mathbf{R}) $ and the final reconstructed image by $ \hat{\mathbf{I}} = \hat{D}(\hat{f}) $.

To enhance reconstruction fidelity, a loss function was introduced to enforce both pixel-level and perceptual constraints while ensuring feature consistency in non-masked regions across different scales. Overall loss function was defined as:
\begin{equation}
\footnotesize
    \mathcal{L} = \|\mathbf{I} - \hat{\mathbf{I}}\|_2 + \sum_{k=1}^{K} \| (1 - M_k) \odot (R_k - \hat{R}_k) \|_2 + \|f - \hat{f}\|_2 + \lambda_P \mathcal{L}_P(\hat{\mathbf{I}}) + \lambda_G \mathcal{L}_G(\hat{\mathbf{I}}),
\end{equation}

\noindent where $M_k \in \{0,1\}^{h_k \times w_k}$ was the lesion mask at scale $k$. The second term was lesion-focused loss, enforcing similarity between real and reconstructed feature tokens in non-masked areas across all scales. $\mathcal{L}_P$ represented a perceptual loss, while $\mathcal{L}_G$ represented a discriminative loss. $\lambda_P$ and $\lambda_G$ were the loss weights.

\subsection{Lesion Measurement Embedded VAR Transformer}

In clinical dermatology, quantified lesion measurements—structured scores that captured morphological characteristics such as lesion texture and shape—were essential for both diagnosis and prognosis \cite{STAAL202152,liu,li}. These structured scores, derived from mask regions rather than full images, offered robust feature representations. Despite their clinical importance, the potential of these structured measurements to guide medical image synthesis remained largely unexplored.

To bridge this gap, we incorporated lesion measurement scores as conditioning signals for controlled skin lesion synthesis. Specifically, during intra-class synthesis, lesion measurement scores $\varepsilon_{\text{r}}$ were extracted from the original images using PyRadiomics \cite{pyradiomics} formulated as $\varepsilon_{\text{r}} = E_{\text{ext}}(\mathbf{I,M})$. These measurement scores included shape, histogram, texture, and clinical attributes. 

To enable autoregressive modelling of discrete tokens, a VAR Transformer was employed, conditioning synthesis on learned embeddings. Along with extracted multi-scale tokens, measurement scores were incorporated via an encoding function, $\mathbf{F}_{\text{q}} = E_{\text{q}}(\varepsilon_{\text{r}} )$. The measurement scores $\varepsilon_{\text{r}}$ were transformed via a linear projection, followed by layer normalization and a SiLU activation function.

For inter-class synthesis, an additional codebook of class-average measurements was maintained. During the intra-class synthesis, $\varepsilon_{\text{r}}$ was used to update the corresponding skin lesion entry in the codebook by computing the average measurements. During the inter-class synthesis, the query feature $\hat{\mathbf{F}_{\text{q}}}$ was obtained by referencing this codebook.

A class embedding $\mathbf{S}$ was introduced to incorporate class-specific information for the generative process. The combined token sequence was defined as $\mathbf{T}_0 = [ \mathbf{S}, \mathbf{F}_{\text{r}} ] $. The VAR Transformer autoregressively predicted the subsequent tokens, denoted as $\hat{\mathbf{R}}$, based on the input sequence $\mathbf{T}_0$ and the reference sequence $\mathbf{R}$. A “next-token prediction” generated the higher-scale skin lesion representations based on the combined tokens and the corresponding lower-scale tokens. The training objective was a standard cross-entropy loss over the token space. 

\section{Experiments and Results}
\subsection{Experiments Setup}

\noindent\textbf{Dataset.} In this study, the HAM10000 \cite{HAM10000} dataset was utilized to train the \textit{LF-VAR} model. The dataset consisted of dermatoscopy images annotated with seven skin lesion categories: actinic keratoses (AKIEC, 327 images), basal cell carcinoma (BCC, 514), benign keratosis (BKL, 1,099), dermatofibroma (DF, 118), melanoma (MEL, 1,113), melanocytic nevi (NV, 6,705), and vascular lesions (VASC, 142). All images were resized to a resolution of 512 × 512 pixels and split into training and testing sets with a 4:1 ratio. For cross-dataset performance evaluation, images from the ISIC2017 \cite{ISIC2017} and Dermofit \cite{Dermofit} datasets that matched the lesion categories in the HAM10000 dataset were selected. 

\noindent \textbf{Baselines.} Several generative methods were selected as baselines, including MAGE \cite{MAGE}, MAGE Adapter \cite{MAGE_Adapter}, Derm T2IM \cite{Derm_T2IM}, VAR \cite{VAR}, and Diffusion \cite{Diffusion}. MAGE and MAGE Adapter were unconditional image generation models. Derm T2IM and VAR utilized text as prompts to guide image synthesis. The Diffusion and VAR model possessed inpainting capabilities, allowing it to generate content within a specified masked region while completing the missing area.

\begin{figure}[t]
\centering
\includegraphics[width= \textwidth]{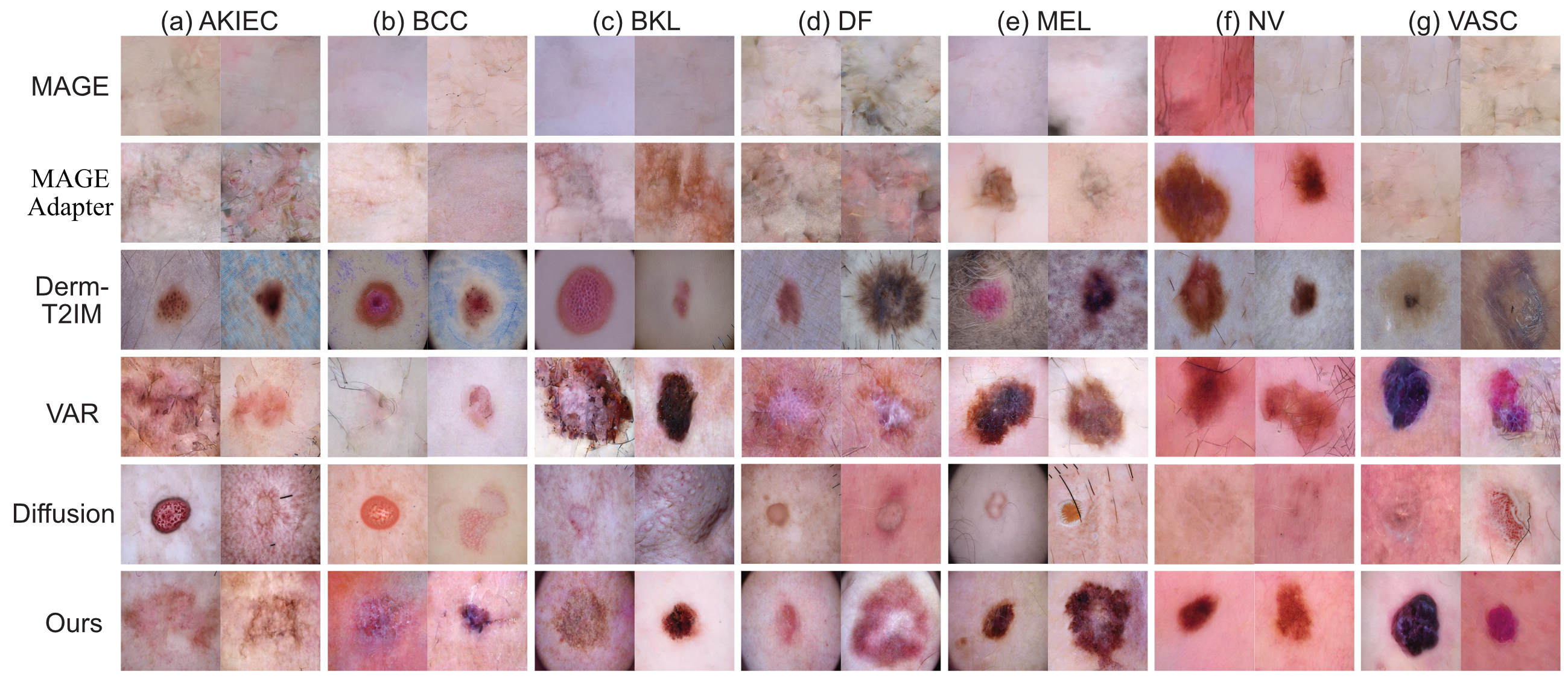}
\caption{Comparison of synthesis samples by baseline and our method, with two samples from different lesions of each model.} \label{result}
\end{figure}

\begin{figure}[t]
\centering
\includegraphics[width=\textwidth]{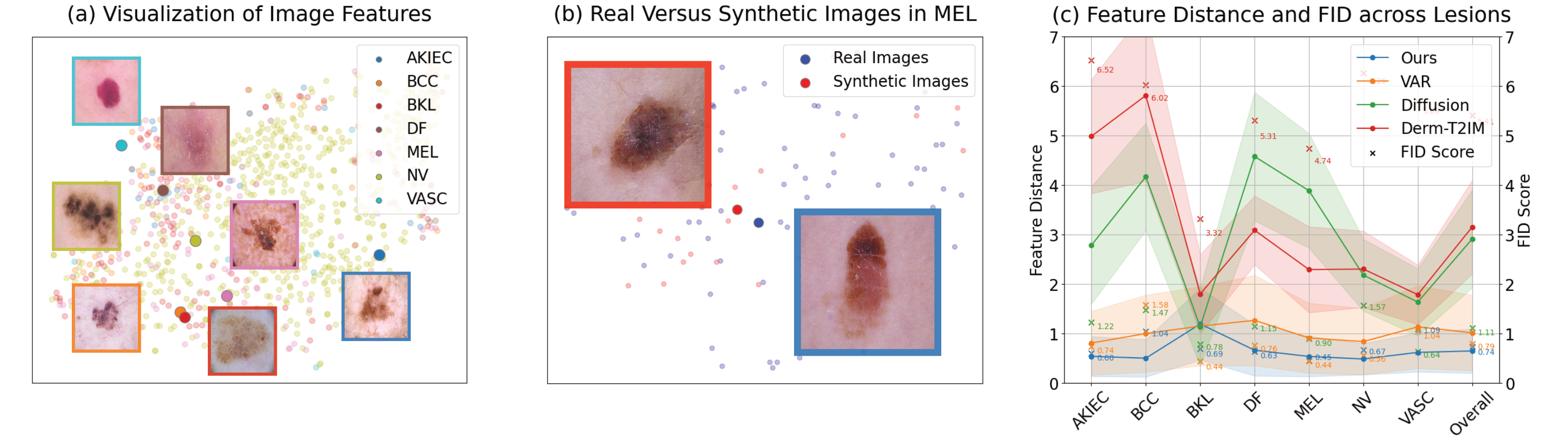}
\caption{Feature analysis and model comparison. (a) Feature distribution across seven skin lesion types. (b) Feature distribution comparison between real and synthetic in melanoma. (c) Synthesis images feature distance and FID comparison among models.}  \label{more_compare}
\end{figure}

\noindent \textbf{Task and metric.} The proposed method was evaluated on three tasks: (1) Intra-class synthesis, to assess the model’s ability to replicate patterns from the training set; (2) Inter-class synthesis, to evaluate its generalizability across different skin lesion types; (3) Cross-dataset evaluation, to test the model’s robustness and generalization on external datasets, and (4) Validation on the downstream classification tasks.

Two metrics were employed: (1) Fréchet Inception Distance (FID) \cite{fid}, which measured the distance between the feature distributions of real and synthesised images, with a lower FID indicating higher fidelity; (2) Inception Score (IS) \cite{is}, which evaluated the quality and diversity of the generated images, with a higher IS reflecting improved variation. In this study, we primarily focus on FID as it provides a more reliable measure of fidelity by comparing feature distributions between real and synthesized images.

\noindent \textbf{Implementation details.} The models were trained using the AdamW optimizer ($\beta_1 = 0.9$, $\beta_2 = 0.95$), a weight decay of 0.05, a learning rate of $1 \times 10^{-3}$, and a batch size of 35 for 200 epochs. All experiments ran on an NVIDIA A100 GPU (40 GB) in a high-performance computing environment. We employed a multi-scale quantization autoencoder\cite{VAR,bai2023sequentialmodelingenablesscalable} and GPT-2 decoder-only transformer \cite{esser2021tamingtransformershighresolutionimage,radford2019language} which keeps consistent with VAR.

\subsection{Results}
\textbf{Intra-class synthesis.} Fig.\ref{result} showed randomly selected skin images generated by different methods for visual comparison. MAGE \cite{MAGE} and MAGE Adapter \cite{MAGE_Adapter} often produced blurred or distorted images with lower visual quality. Derm T2IM \cite{Derm_T2IM} and Diffusion \cite{Diffusion} models generated higher-quality results, though inconsistencies were noticeable, especially for melanoma. VAR \cite{VAR} produced visually realistic images but failed to capture key lesion characteristics, with inconsistent lesion colour, shape, and unnatural transitions. In contrast, our method improved skin synthesis quality through lesion measurement scores and reduced artifacts with a well-designed multi-scale lesion-focused VQ-VAE, resulting in more realistic skin lesion images.

Fig.\ref{more_compare} presented a feature space analysis of real and generated images. The t-distributed Stochastic Neighbor Embedding (t-SNE) \cite{t-sne} visualization showed distinct clusters for each skin lesion type. Feature distance comparisons and FID scores across lesion types indicated that our method achieved better feature consistency and synthesis quality compared to baseline models. Table \ref{performance_test} presented the quantitative evaluation of synthesis performance using FID and IS metrics. Our method achieved the best overall FID score (average 0.74), improving upon the previous state-of-the-art (SOTA) by 6.3\%. Derm T2IM showed high IS across most categories (4.28), due to the introduction of artifacts into normal skin regions, which increased diversity and inflated the IS metric. In contrast, our method maintained an IS of 2.95 while preserving lesion fidelity and minimizing artifacts, as reflected in its lowest FID scores.

\begin{figure}[t]
\centering
\includegraphics[width=\textwidth]{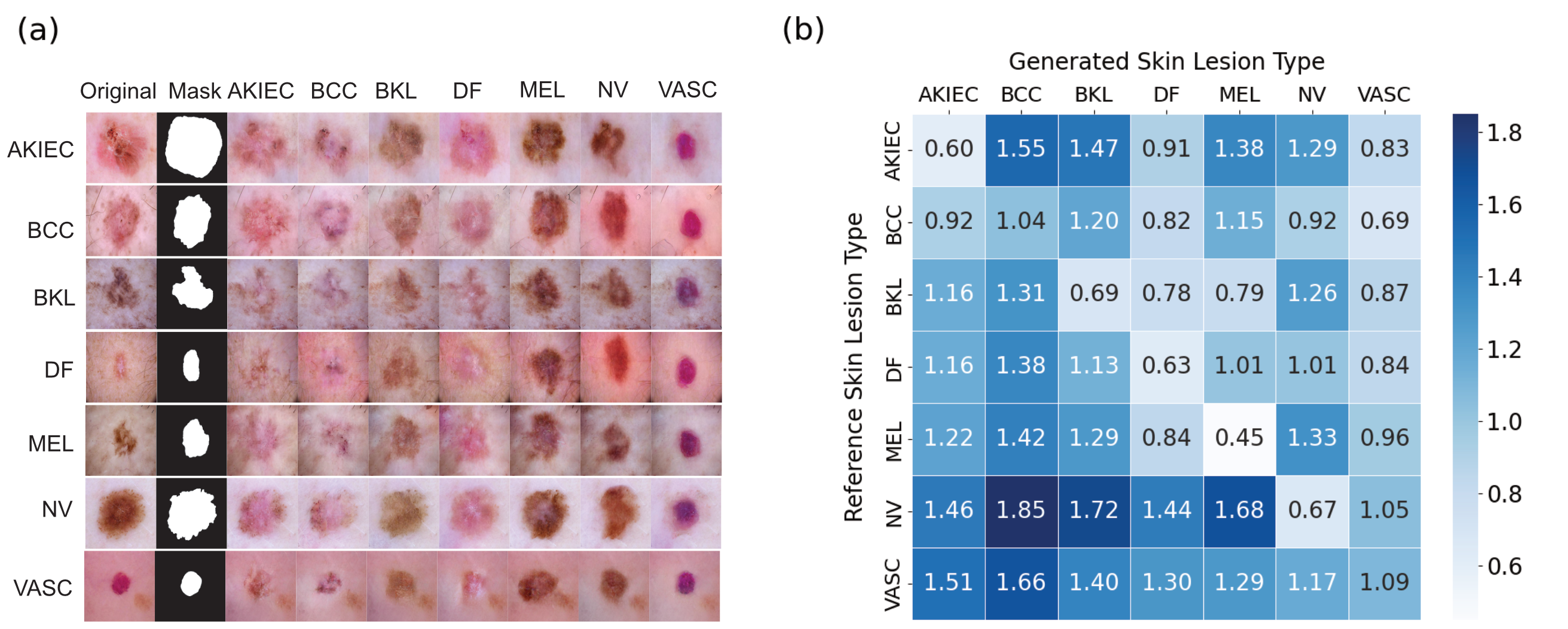}
\caption{Inter-class synthesis comparison and FID matrix. (a) Inter-class synthesis samples for seven lesion types. The first two columns show the original images and masks, followed by synthesis images. (b) FID confusion matrix for inter-class synthesis.}\label{conf_matrix}
\end{figure}

\noindent\textbf{Inter-class synthesis.} We evaluated the inter-class synthesis performance of our model using different text-based lesion categories as prompts. The results were presented in Fig.~\ref{conf_matrix}. As shown in the confusion matrix, all lesion categories achieved the best synthesis performance when generating VASC, with an average FID of 0.90 (standard deviation [std], 0.13), followed by DF (mean, 0.96; std, 0.27). The comparative figure further showed that our model preserved a clear background and smooth lesion boundaries while maintaining consistent lesion characteristics across columns, ensuring the capture of distinct lesion features in inter-class synthesis.

\begin{table}[t]
\caption{Performance comparison on the test dataset. \textbf{Bold} indicates the best FID, while \underline{underlined} indicates the second-best.}
\label{performance_test}
\centering
\resizebox{0.8\linewidth}{!}{
\begin{tabular}{l|c|l|rrrrrrr|r}
\toprule
Method  & Prompt       & Matrix &  AKIEC &  BCC &  BKL & DF & MEL &  NV &  VASC & Average \\ 
\midrule
MAGE \cite{MAGE}   & \ding{55}  & IS ↑   & \makecell[r]{3.08 \\ ± 0.13} & \makecell[r]{3.18 \\ ± 0.12} & \makecell[r]{3.01 \\ ± 0.15} & \makecell[r]{2.96 \\ ± 0.21} & \makecell[r]{2.91 \\ ± 0.16} & \makecell[r]{2.98 \\ ± 0.14} & \makecell[r]{3.08 \\ ± 0.11} & \makecell[r]{3.03 \\ ± 0.15} \\
   & & FID ↓  & 12.92  & 11.86 & 9.84  & 12.18 & 8.49  & 14.97 & 16.34  & 12.37  \\ 
\midrule
MAGE Adapter \cite{MAGE_Adapter} & \ding{55} & IS ↑   & \makecell[r]{2.49 \\ ± 0.05} & \makecell[r]{1.79 \\ ± 0.09} & \makecell[r]{1.98 \\ ± 0.05} & \makecell[r]{2.84 \\ ± 0.14} & \makecell[r]{2.38 \\ ± 0.09} & \makecell[r]{2.66 \\ ± 0.10} & \makecell[r]{2.67 \\ ± 0.16} & \makecell[r]{2.40 \\ ± 0.10} \\
           &   & FID ↓  & 2.33  & 7.99 & 2.65 & 3.09  & 5.82 & 3.24  & 3.28  & 4.06 \\ 
\midrule
Derm T2IM \cite{Derm_T2IM} & Text & IS ↑   & \makecell[r]{3.74 \\ ± 0.19} & \makecell[r]{4.23 \\ ± 0.17} & \makecell[r]{3.94 \\ ± 0.39} & \makecell[r]{5.03 \\ ± 0.31} & \makecell[r]{3.59 \\ ± 0.29} & \makecell[r]{4.74 \\ ± 0.30} & \makecell[r]{4.69 \\ ± 0.46} & \makecell[r]{4.28 \\ ± 0.30} \\
           &   & FID ↓  & 6.52  & 6.02  & 3.32 & 5.31  & 4.74  & 6.26  & 5.69  & 5.41  \\ 
\midrule
Diffusion \cite{Diffusion} & Mask & IS ↑   & \makecell[r]{3.90 \\ ± 0.20} & \makecell[r]{4.88 \\ ± 0.25} & \makecell[r]{3.12 \\ ± 0.12} & \makecell[r]{4.04 \\ ± 0.18} & \makecell[r]{4.22 \\ ± 0.30} & \makecell[r]{3.96 \\ ± 0.25} & \makecell[r]{3.10 \\ ± 0.09} & \makecell[r]{3.89 \\ ± 0.20} \\
            &  & FID ↓  & 1.22  & \underline{1.47}  & 0.78  & 1.15  & 0.90  & 1.57  & \textbf{0.64}  & 1.11  \\ 
\midrule
VAR \cite{VAR} & Text- & IS ↑   & \makecell[r]{3.09 \\ ± 0.14} & \makecell[r]{2.57 \\ ± 0.12} & \makecell[r]{3.27 \\ ± 0.19} & \makecell[r]{2.58 \\ ± 0.08} & \makecell[r]{2.76 \\ ± 0.11} & \makecell[r]{2.37 \\ ± 0.10} & \makecell[r]{2.48 \\ ± 0.15} & \makecell[r]{2.73 \\ ± 0.13} \\
           &  Mask & FID ↓  & \underline{0.74} & 1.58  & \textbf{0.44}  & \underline{0.76}  & \textbf{0.44}  & \textbf{0.56}  & \underline{1.04}  & \underline{0.79}  \\ 
\midrule
Ours & Text-   & IS ↑   & \makecell[r]{3.27 \\ ± 0.13} & \makecell[r]{2.41 \\ ± 0.09} & \makecell[r]{3.26 \\ ± 0.12} & \makecell[r]{2.34 \\ ± 0.11} & \makecell[r]{2.84 \\ ± 0.08}  & \makecell[r]{2.63 \\ ± 0.13} & \makecell[r]{3.92 \\ ± 0.08} & \makecell[r]{2.95 \\ ± 0.10}  \\
       & Mask    & FID ↓  &  \textbf{0.60}  & \textbf{1.04}  & \underline{0.69}  & \textbf{0.63}  & \underline{0.45}  & \underline{0.67}  & 1.09  & \textbf{0.74}  \\ 
\bottomrule
\end{tabular}}
\end{table}

\noindent\textbf{Cross-dataset synthesis.} Table \ref{performance_cross} presented the evaluation results on two external datasets. Our method achieved the best average performance across both datasets. Specifically, on the ISIC2017 dataset, it obtained the lowest FID (1.70), outperforming all other methods across most categories, particularly in BKL (1.94) and MEL (1.99). On the Dermofit dataset, our method achieved the best average FID (1.71), with the lowest FID in AKIEC (1.29), BCC (0.56), and VASC (0.38). These results highlighted the synthesis capability of our model compared to existing approaches, showing consistent performance across diverse datasets.

\noindent\textbf{Downstream task validation.} Following the established protocol by Philipp et al. \cite{tschandl2020human}, our downstream classification uses a ResNet-50 classifier for 7 lesion types. The baseline achieved a mean recall of 0.692. Adding a weighted random sampler improved it to 0.715. When using augmented training data with our synthetic images (balanced to 500 images per class), the recall increased to 0.771 (+0.079, +11.4\% from baseline; +0.056, +7.8\% from weighted sampling). This demonstrates the practical utility of our synthetic data for improving classification performance.

\begin{table}[t]
\caption{FID performance comparison on cross-dataset synthesis.}\label{performance_cross}
\centering
\resizebox{0.8\linewidth}{!}{ 
\begin{tabular}{l|l|rrrrrrr|r}
\toprule
Dataset  & Method       & AKIEC & BCC & BKL & DF & MEL & NV & VASC & Average \\ 
\midrule
ISIC2017 \cite{ISIC2017} 
         & MAGE \cite{MAGE}            &   -    &   -  &   \makecell[r]{8.04}  &  -  &    \makecell[r]{6.99}  &  \makecell[r]{8.34}  &  -    &    \makecell[r]{7.94}     \\
         & MAGE Adapter \cite{MAGE_Adapter}  &   -    &   -  &  \makecell[r]{5.89}   &  -  &  \makecell[r]{8.19}   & \makecell[r]{9.75}   &  -    &  \makecell[r]{7.79}       \\
         & Derm T2IM  \cite{Derm_T2IM}  &   -    &   -  & \makecell[r]{4.22}    &  -  &  \makecell[r]{6.01}   & \makecell[r]{\underline{3.88}}   &  -    &  \makecell[r]{4.71}       \\
         & VAR  \cite{VAR} &   -    &   -  &  \makecell[r]{2.49}   &  -  &   \makecell[r]{\underline{2.13}}  &  \makecell[r]{4.21}  &  -    &    \makecell[r]{\underline{2.94}}      \\
         & Diffusion \cite{Diffusion}&   -    &   -  &  \makecell[r]{\underline{2.36}}   &  -  &    \makecell[r]{7.20} &  \makecell[r]{12.98}  &  -    &  \makecell[r]{7.51}       \\
         & Ours  &   -    &   -  &  \makecell[r]{\textbf{1.94}}   &  -  &  \makecell[r]{\textbf{1.99}}   &  \makecell[r]{\textbf{1.18}}  &  -    & \makecell[r]{\textbf{1.70}}       \\ 
\midrule
Dermofit \cite{Dermofit} 
         & MAGE \cite{MAGE}        &  \makecell[r]{11.01}    &  \makecell[r]{11.77}   &  \makecell[r]{10.42}   &  \makecell[r]{7.80}   &  \makecell[r]{11.32}   &  \makecell[r]{9.00}   &  \makecell[r]{7.91}    &  \makecell[r]{9.89}       \\
         & MAGE Adapter \cite{MAGE_Adapter}  &  \makecell[r]{6.88}     & \makecell[r]{16.02}    &  \makecell[r]{8.76}    &  \makecell[r]{5.31}   & \makecell[r]{8.24}    &   \makecell[r]{15.90}  &  \makecell[r]{6.89}    &      \makecell[r]{9.71}   \\
         & Derm T2IM \cite{Derm_T2IM}   &   \makecell[r]{8.29}     & \makecell[r]{10.69}    &  \makecell[r]{9.08}   & \makecell[r]{9.20}   & \makecell[r]{13.48}    &  \makecell[r]{6.78}   &  \makecell[r]{7.27}    &  \makecell[r]{9.26}       \\
         & VAR \cite{VAR}         &    \makecell[r]{4.94}    &   \makecell[r]{5.40}   &  \makecell[r]{4.98}    &   \makecell[r]{5.78}   &    \makecell[r]{4.64}  &   \makecell[r]{8.25}  &  \makecell[r]{9.43}    &        \makecell[r]{6.20}   \\
         & Diffusion \cite{Diffusion}    &   \makecell[r]{\underline{2.20}}     &   \makecell[r]{\underline{3.88}}   &  \makecell[r]{\textbf{2.23}}   & \makecell[r]{\textbf{2.67}}   &  \makecell[r]{12.73}   &   \makecell[r]{\textbf{2.38}}  &  \makecell[r]{\underline{1.90}}    & \makecell[r]{\underline{4.00}}         \\
         & Ours         &   \makecell[r]{\textbf{1.29}}    &  \makecell[r]{\textbf{0.56}}   &   \makecell[r]{\textbf{2.23}}   &   \makecell[r]{\underline{3.92}}  &  \makecell[r]{\textbf{0.39}}    &   \makecell[r]{\underline{3.23}}  &   \makecell[r]{\textbf{0.38}}   &  \makecell[r]{\textbf{1.71}}       \\ 
\bottomrule
\end{tabular}}
\end{table}

\noindent\textbf{Ablation study.} To validate the effectiveness of lesion focus VQ-VAE and measurement embedding in enhancing VAR, we systematically integrated specific modules into the baseline. We sequentially introduced the following components: \textbf{LF}: a multi-scale lesion focus VQ-VAE; \textbf{FM}: a fixed measurement embedding where a constant measurement value was applied across all lesions; and \textbf{AM}: an adaptive measurement embedding that maintained a class-specific average measurement codebook and dynamically assigned measurements based on lesion type.

Table \ref{ablation} summarised the experimental study results, demonstrating a progressive improvement in synthesis performance with the incorporation of different modules. Integrating LF into the baseline led to an average performance gain of 3.7\%, while combining LF and AM led to the best overall performance, achieving the lowest average FID (0.74) and the highest IS (2.95 ± 0.10). Compared to the baseline, combining LF and AM improved performance by 6.3\%, surpassing the baseline with LF by 2.6\%.

\begin{table}[t]
\caption{Ablation study results. \textbf{Bold} indicates the best FID, while \underline{underlined} represents the second-best.}\label{ablation}
\centering
\resizebox{0.8\linewidth}{!}{ % Adjust width
\begin{tabular}{l|l|rrrrrrr|r}
\toprule
    Setting & Metric &  AKIEC  &  BCC   &  BKL   & DF   & MEL  &  NV   &  VASC  & Average   \\ 
\midrule
Baseline & IS ↑   & \makecell[r]{3.09\\ ± 0.14} & \makecell[r]{2.57\\ ± 0.12} & \makecell[r]{3.27\\ ± 0.19} & \makecell[r]{2.58\\ ± 0.08} & \makecell[r]{2.76\\ ± 0.11} & \makecell[r]{2.37\\ ± 0.10} & \makecell[r]{2.48\\ ± 0.15} & \makecell[r]{2.73\\ ± 0.13} \\
         & FID ↓  & 0.74 & 1.58 & \textbf{0.44} & 0.76 & \textbf{0.44} & \textbf{0.56} & 1.04 & 0.79 \\ 
\midrule
Baseline + LF & IS ↑   & \makecell[r]{3.42\\ ± 0.10} & \makecell[r]{2.52\\ ± 0.10} & \makecell[r]{3.47\\ ± 0.26} & \makecell[r]{3.06\\ ± 0.11} & \makecell[r]{3.01\\ ± 0.12} & \makecell[r]{2.67\\ ± 0.17} & \makecell[r]{2.35\\ ± 0.08} & \makecell[r]{2.93\\ ± 0.13} \\
            & FID ↓  & \underline{0.70} & \underline{1.23} & \underline{0.47} & \textbf{0.62} & 0.54 & 0.74 & \underline{1.00} & \underline{0.76} \\ 
\midrule
Baseline + LF + FM & IS ↑   & \makecell[r]{2.69\\ ± 0.08} & \makecell[r]{2.05\\ ± 0.07} & \makecell[r]{2.95\\ ± 0.17} & \makecell[r]{2.65\\ ± 0.11} & \makecell[r]{2.42\\ ± 0.09} & \makecell[r]{2.25\\ ± 0.12} & \makecell[r]{1.77\\ ± 0.03} & \makecell[r]{2.40\\ ± 0.10} \\
              & FID ↓  & 1.14 & 2.09 & 0.73 & 0.72 & 1.16 & \underline{0.62} & \textbf{0.91} & 1.05 \\ 
\midrule
Baseline + LF + AM  & IS ↑   & \makecell[r]{3.27\\ ± 0.13} & \makecell[r]{2.41\\ ± 0.09} & \makecell[r]{3.26\\ ± 0.12} & \makecell[r]{2.34\\ ± 0.11} & \makecell[r]{2.84\\ ± 0.08} & \makecell[r]{2.63\\ ± 0.13} & \makecell[r]{3.92\\ ± 0.08} & \makecell[r]{2.95\\ ± 0.10} \\
            & FID ↓  & \textbf{0.60} & \textbf{1.04} & 0.69 & \underline{0.63} & \underline{0.45} & 0.67 & 1.09 & \textbf{0.74} \\ 
\bottomrule
\end{tabular}}
\end{table}

\section{Conclusion}
In this paper, we introduced \textit{LF-VAR}, a controllable skin lesion synthesis model that integrates lesion measurement scores with a lesion-focused VQ-VAE and VAR model. By leveraging structured lesion measurements as conditional signals, our approach enables precise control over key lesion characteristics while maintaining high synthesis fidelity. The results demonstrate the effectiveness of integrating clinically relevant features into generative models, highlighting the potential of \textit{LF-VAR} for enhancing skin image synthesis in medical applications.

\begin{credits}

\subsubsection{\discintname}
The authors have no competing interests to declare that are
relevant to the content of this article.
\end{credits}

\bibliographystyle{splncs04}
\bibliography{mybibliography}
\end{document}